# Automatic marker-free registration based on similar tetrahedras for single-tree point clouds

Jing Ren [1], Pei Wang[1*], Hanlong Li[1], Yuhan Wu[1], Yuhang Gao[1], Wenxin Chen[1], Mingtai Zhang[1], Lingyun Zhang[1]

*Abstract*—In recent years, terrestrial laser scanning technology has been widely used to collect tree point cloud data, aiding in measurements of diameter at breast height, biomass, and other forestry survey data. Since a single scan from terrestrial laser systems captures data from only one angle, multiple scans must be registered and fused to obtain complete tree point cloud data. This paper proposes a marker-free automatic registration method for single-tree point clouds based on similar tetrahedras. First, two point clouds from two scans of the same tree are used to generate tree skeletons, and key point sets are constructed from these skeletons. Tetrahedra are then filtered and matched according to similarity principles, with the vertices of these two matched tetrahedras selected as matching point pairs, thus completing the coarse registration of the point clouds from the two scans. Subsequently, the ICP method is applied to the coarse-registered leaf point clouds to obtain fine registration parameters, completing the precise registration of the two tree point clouds. Experiments were conducted using terrestrial laser scanning data from eight trees, each from different species and with varying shapes. The proposed method was evaluated using RMSE and Hausdorff distance, compared against the traditional ICP and NDT methods. The experimental results demonstrate that the proposed method significantly outperforms both ICP and NDT in registration accuracy, achieving speeds up to 593 times and 113 times faster than ICP and NDT, respectively. In summary, the proposed method shows good robustness in single-tree point cloud registration, with significant advantages in accuracy and speed compared to traditional ICP and NDT methods, indicating excellent application prospects in practical registration scenarios.

## I. INTRODUCTION

In recent years, terrestrial laser scanning (TLS) has gradually become an important technological tool for forestry surveys and management [1]. It is commonly used to obtain detailed 3D structural information of trees, supporting the estimation of various forest survey parameters such as tree height, individual tree biomass [2][3], volume [4][5], trunk morphology [6][7][8], and leaf area [9][10]. However, in forest environments, the spatial coverage of a single TLS scan is limited, and nearby trees and shrubs often block the laser beam, affecting the scanning of more distant trees. Studies have shown that in a single TLS scan, 10%-32% of trees may not be scanned due to their position [11]. Even for nearby trees, a single TLS scan only captures partial information of the trees. Therefore, to obtain complete tree structures for subsequent analysis and estimation, multiple scans from different viewpoints are needed, and multi-scan point cloud registration must be performed [12].

Multi-scan registration can generally be divided into two types: marker-based registration and marker-free registration. For marker-based registration, physical markers, such as reflective targets, are typically arranged in the scene to assist in aligning point clouds from different viewpoints [13]. Since markers provide explicit reference points, this method allows for rapid matching of multi-scan point clouds and reduces the computational complexity of feature extraction. Such methods usually also achieve good registration accuracy. However, the setup of markers requires extra time, which is particularly challenging for large-scale or complex outdoor environments [14].

[1] *School of Science, Beijing Forestry University, Beijing 100083, China*
*Corresponding author: email: wangpei@bjfu.edu.cn.

Marker-free registration can also be divided into two categories. The first category directly uses all points for global registration, such as the Iterative Closest Point (ICP) method [15]. The second category extracts natural feature points from the point cloud based on feature descriptors for registration, such as Fast Point Feature Histograms (FPFH) and Signature of Histograms of Orientations (SHOT) [16][17][18]. For the ICP method, the computational speed depends on the initial alignment and the size of the point cloud; if the initial alignment is good and the point cloud is small, it can converge quickly with high efficiency and accuracy; otherwise, the opposite is true [19]. As for the second category of methods, they can extract feature points effectively in feature-rich environments, achieving good registration accuracy. However, in feature-poor or noisy environments, the accuracy of registration is easily affected, although these methods typically have good efficiency [20].

In forest environments, registration methods often use the positional relationship of tree trunks to aid the process. Liang et al. proposed a method that independently processes each scan and registers the data at the feature and decision levels without relying on artificial references, resulting in high processing efficiency. However, this registration method is also affected by occlusions and incomplete point cloud data [21]. Liu et al. proposed another automatic registration method without markers, which utilizes the natural geometric features of tree trunks and registers based on the morphological profiles of trunks at different heights. However, when the morphological characteristics of trees in the forest environment are similar, it affects the mapping and registration accuracy [22]. Kelbe et al. proposed a method that does not rely on the initial sensor pose estimation, but instead generates viewpoint-invariant feature descriptors and uses a voting algorithm to determine the optimal registration parameters automatically, achieving blind registration of forest point clouds. This method demonstrates good robustness in environments with dense understory vegetation and significant viewpoint changes, but is highly sensitive to feature point selection, which can lead to registration failure if the feature points are not appropriately selected [23]. Ge et al. proposed an innovative strategy based on common subgraphs, which first extracts initial feature nodes through a model-driven approach and then uses graph theory methods for network optimization, significantly improving registration accuracy. However, this method has high computational complexity and consumes significant resources when dealing with large-scale forest point cloud data [24]. Wang et al. proposed a global registration method based on the relative positions of tree trunks, which utilizes the relative spatial relationships of trunks from different scanning angles to determine registration transformation parameters and then achieves accurate registration through a non-iterative process using local triangle construction and a global matching strategy. However, when the spatial distribution of trees is relatively close, registration errors may occur [25].

Compared to forest scene point clouds, single-tree point clouds lack the relative positional relationships between trees. Without markers, and considering the complexity of tree structures, such as complex branch structures, diverse growth forms, and the incomplete data caused by occlusions during scanning, achieving automated, accurate marker-free registration for single-tree point clouds presents significant challenges. Bucksch and Khoshelham proposed a local registration method in 2013 based on skeletonization, which registers point clouds by detecting corresponding branches from multiple scans. The accuracy of coarse registration significantly impacts the final fine registration [26]. Similarly, Zhou et al. proposed an automatic coarse registration method based on skeleton extraction, using the root position, branch lengths, and skeleton correspondences for accurate registration [27].

To address these challenges, this paper proposes a marker-free automatic registration method for single-tree point clouds based on similar tetrahedra (AMRST, Automatic Marker-free Registration based on Similar Tetrahedras). In this method, we employ branch-leaf separation, skeleton generation, and extraction of key similar tetrahedra pairs, significantly improving the accuracy, efficiency, and robustness of single-tree point cloud registration. The structure of this paper is as follows: Section 2 details the workflow of the AMRST method; Section 3 verifies the registration accuracy and efficiency of AMRST based on experimental data and compares it with traditional methods; Sections 4 and 5 discuss and conclude the characteristics of the AMRST method and future research directions.

## II. METHODS

### A. Data

The equipment used in the experiment was the RIEGL VZ-400 terrestrial laser scanner (RIEGL Laser Measurement Systems GmbH, 3580 Horn, Austria), with its specifications listed in Table 1. The angular resolution used for data acquisition was 0.02 degrees. The registration method presented in this paper was developed using the C++ programming language, based on the Point Cloud Library (PCL). The experiment was conducted on a computer equipped with a 2.9 GHz Intel Core i7 eight-core processor and 32 GB of RAM.

TABLE I
CHARACTERISTICS OF RIEGL VZ – 400

| Technical parameters | |
|---|---|
| The farthest distance measurement | 600 m (natural object reflectivity≥90%) |
| The scanning rate (points / second) | 300,000 (emission), 125,000 (reception) |
| The scanning range | −40° ~ 60°(vertical) 0° ~ 360°(horizontal) |
| Laser divergence | 0.3 mrad |
| Connection | LAN / WLAN, wireless data transmission |

The tree point cloud data used in the experiment was collected in May 2021 at Bajia Park, Haidian District, Beijing, and in September 2023 at Beijing Forestry University, Haidian District, Beijing. A total of eight trees belonging to different species were scanned, with two scans per tree. The trees exhibited significant morphological differences. The trees were labeled and sorted as follows: *Sophora japonica,* *Prunus persica, Prunus armeniaca, Koelreuteria paniculata, Pistacia chinensis, Populus spp., Ulmus spp., and Betula spp.*

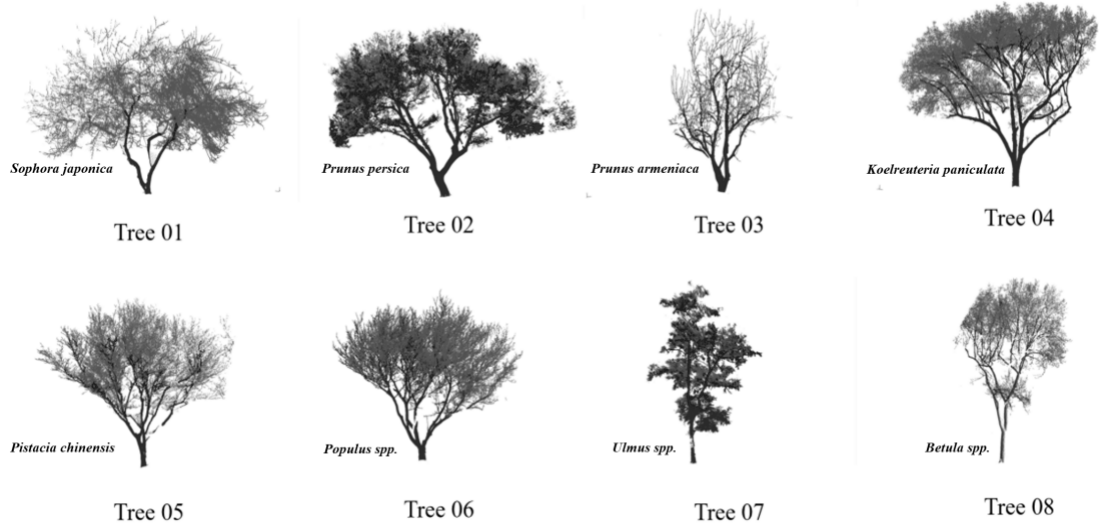

**Fig. 1.** Demonstration of eight tree morphology

*B. Experimental Methods*

The detailed workflow of the proposed tree point cloud registration method is shown in Figure 2. First, the original tree point cloud is subjected to wood-leaf separation, dividing it into branch point clouds and leaf point clouds. Using the separated branch point clouds, a skeleton point cloud of the tree is constructed, capturing the core structural features of the tree. Next, key branch points and end points are identified in the skeleton point cloud, and an initial transformation matrix between the two skeleton point clouds is calculated based on these key points to achieve coarse registration. Then, the LM-ICP (Levenberg-Marquardt Iterative Closest Point) algorithm is applied to the leaf point clouds after coarse registration for fine registration, obtaining the fine registration parameters and completing the registration of the entire tree point cloud dataset. Finally, the registration results are evaluated using RMSE and Hausdorff distance.

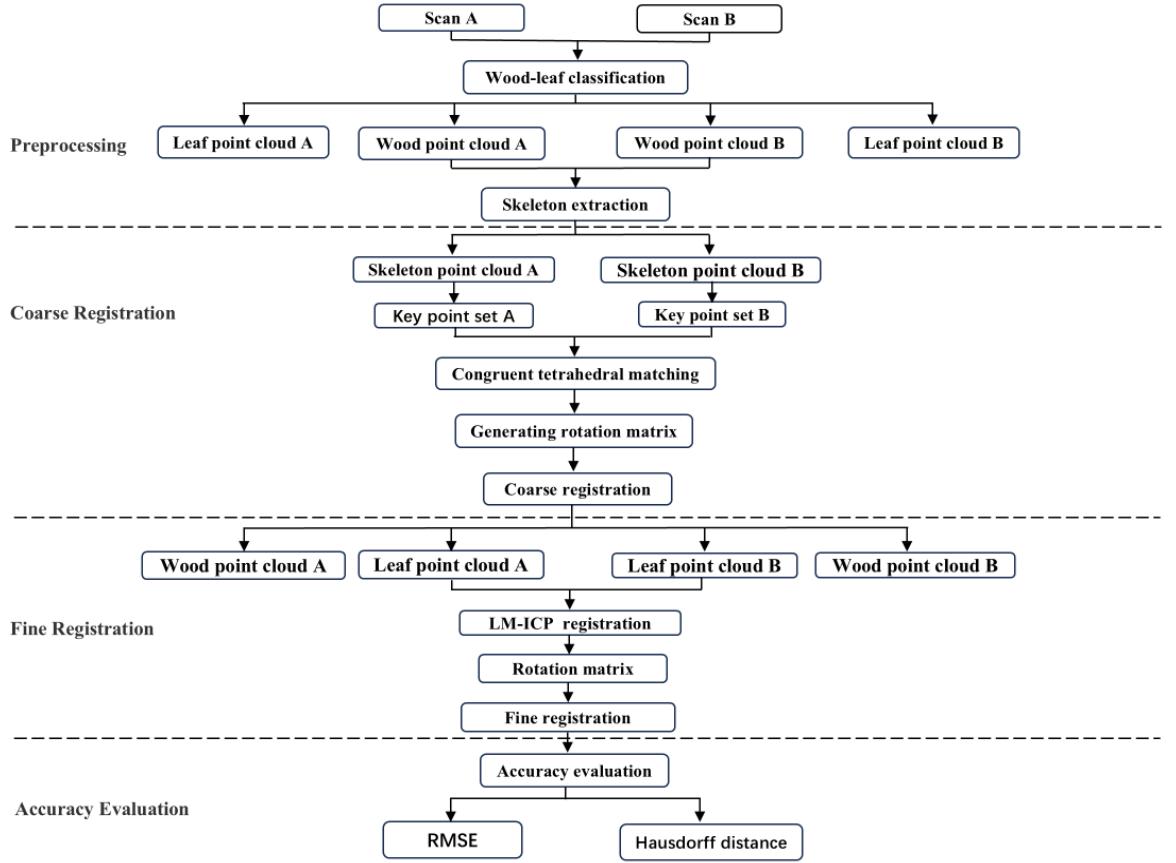

**Fig. 2.** Flowchart of the AMRST method

*C. Preprocessing*

The preprocessing of the AMRST method includes noise filtering, wood-leaf separation, and tree skeleton extraction. Through data preprocessing, environmental and measurement noise in the point cloud data can be reduced, while the branch point cloud obtained from wood-leaf separation is used to generate the tree skeleton for subsequent coarse registration. The wood-leaf separation method used in this paper is based on point cloud intensity and geometric information, using intensity data, K-nearest neighbor algorithm, and voxel processing in sequence to achieve wood-leaf separation. This method is characterized by automation, high speed, and accuracy [34]. Figure 3 shows the wood-leaf separation results for Tree04, where the wood points are displayed in brown and the leaf points in green. Clearly, the separation yields distinct branch and leaf point clouds.

from two scans. It is evident that the generated skeleton effectively reflects the real structure of the tree.

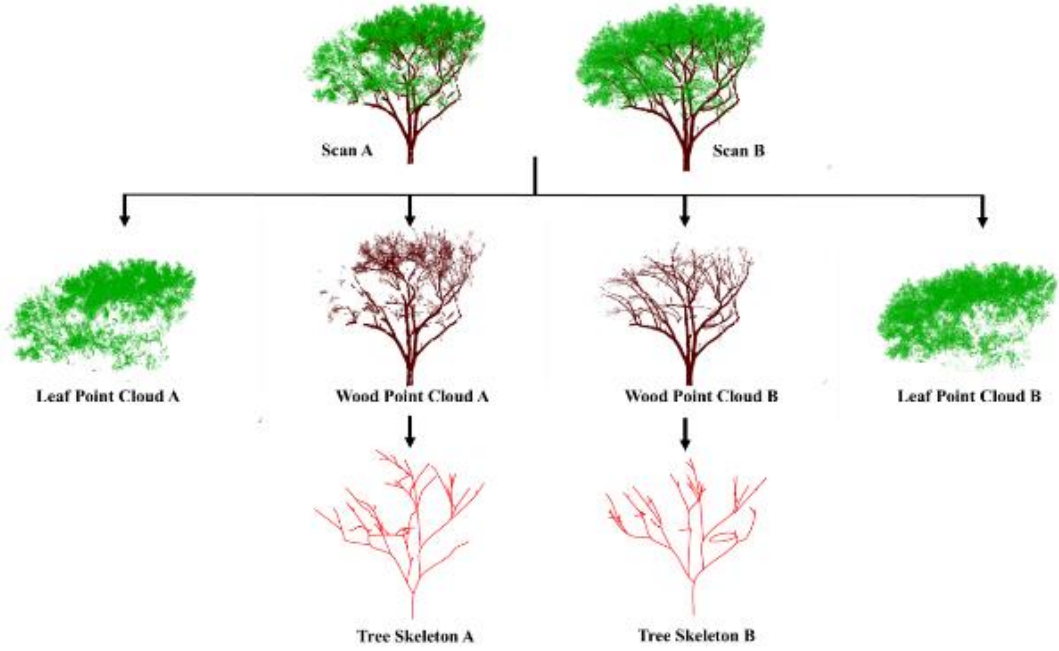

**Fig. 3.** Schematic diagram of preprocessing for two scans (taking Tree4 as an example). Wood points are brown and leaf points are green.

Next, the tree skeleton is extracted based on the Minimum Spanning Tree (MST) algorithm to obtain the branch structure of the tree [35]. The specific steps include: first generating an initial adjacency graph through Delaunay triangulation, then calculating the Minimum Spanning Tree using Dijkstra's shortest path algorithm to obtain the initial skeleton point set of the tree, and finally optimizing the skeleton structure through geometric centralization to reflect the tree's topological morphology. Figure 3 shows the skeleton generation of Tree04 based on the branch point cloud after wood-leaf separation.

*D. Coarse Registration*

The coarse registration in this method mainly includes tree skeleton key point extraction, key point pair selection, and transformation matrix calculation. The detailed steps are as follows.

1) **Key Point Extraction**

Traditional registration methods generally select key points based on target features such as normal vectors and curvature. However, due to the diverse and complex morphology of trees, it is challenging to select consistent key points based on regular positions. The structure of tree branches, as a unique feature of the tree itself, is not affected by the observation angle. Therefore, multi-scan data acquired from different stations can consistently capture the tree skeleton structure with similar features.

Based on the tree skeleton, an undirected graph can be constructed to clarify the connections between skeleton nodes. Each node in this undirected graph corresponds to a node of the tree skeleton, while the edges represent the neighboring or connected relationships between skeleton nodes. If a node has more than two child nodes, it indicates a branching point at that position in the tree, and such nodes are referred to as branch points. Another type of node is called an end point, which includes nodes without any child nodes (representing the ends of branches) or without a parent node (representing the root of the tree). Both branch points and end points in the skeleton are key points that provide a reference for tree point cloud registration.

After node selection is completed, the corresponding nodes are connected step-by-step from the root node of the tree skeleton upwards. Considering the reliability of the trunk and main branch structure, a depth judgment mechanism is introduced, selecting the first five nodes along the path from the root to each end point as key points. Introducing a depth judgment mechanism helps reduce data noise, especially minimizing the impact of small branches that are difficult to accurately distinguish on the skeleton, thereby ensuring the reliability of key points. This also improves the geometric and topological accuracy of the skeleton for subsequent registration, enhancing the robustness of the algorithm.

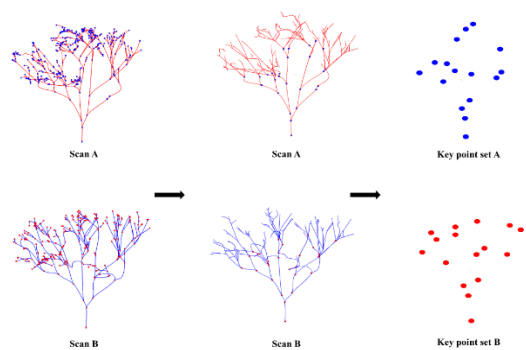

**Fig. 4.** Schematic diagram of keypoint set construction.

Figure 4 illustrates the process of constructing the key point set from the tree skeleton. The left image shows the initially constructed key point set on the tree skeleton, the middle image shows the key point set on the tree skeleton after introducing the depth judgment mechanism, and the right image shows the final filtered key point set. Due to the stable structure of the trunk, the tree point clouds from different stations maintain a high degree of structural consistency despite differences. In particular, the branch points in the branches exhibit high distinguishability in terms of structure.

**2) Key Point Matching**

Based on the constructed key point sets from the skeletons of the two scans, this paper estimates the coarse registration parameters by searching for similar tetrahedra within the two key point sets. The vertices of the two similar tetrahedra are selected from the key point sets of the respective tree skeletons, meaning that four point pairs that can form similar tetrahedra are sought in both key point sets. The tetrahedral structure can be considered as part of the branching structure of the tree, which is commonly present in both skeletons of the two scans. Once four point pairs that meet the requirements are found, the coarse registration of the two tree point clouds can be performed based on these four point pairs.

A tetrahedron is the simplest polyhedron in three-dimensional space, consisting of four points, with each face being a triangle. This structure ensures stability and uniqueness in space. According to the tetrahedral congruence theorem, two tetrahedra are congruent if all corresponding edge lengths are equal [36]. Based on this principle, the following steps are used to achieve tetrahedron matching and key point pair selection:

(1) Calculation of Equal-Length Edges. Let the two sets of key points b $P_1 = \{p_1, p_2, \ldots, p_n\}$ and $P_2 = \{q_1, q_2, \ldots, q_m\}$, where $p_i$ and $q_j$ are points in each key point set. Calculate the Euclidean distances between all points within each set of key points to generate a distance matrix for the point pairs. For any two points $p_i$ and $p_j$ in $P_1$, the calculation is as follows:

$$d_{ij}^{(1)} = |p_i - p_j|$$
$$= \sqrt{(x_i - x_j)^2 + (y_i - y_j)^2 + (z_i - z_j)^2}$$

Similarly, for any two points $q_k$ and $q_l$ in set $P_2$, the distance can be calculated as follows:

$$d_{kl}^{(2)} = |q_k - q_l|$$
$$= \sqrt{(x_k - x_l)^2 + (y_k - y_l)^2 + (z_k - z_l)^2}$$

Next, compare the two sets of distances $\{d_{ij}^{(1)}\}$ and $\{d_{kl}^{(2)}\}$, and filter out edges that meet the following criteria:

$$\frac{|d_{ij}^{(1)} - d_{kl}^{(2)}|}{\max\left(d_{ij}^{(1)}, d_{kl}^{(2)}\right)} < \epsilon$$

Due to spatial errors in generating the tree skeleton and extracting key points, the values describing the same segment in the two point sets may have deviations and are not exactly identical. Therefore, a tolerance $\epsilon$ is introduced in the filtering criteria. If the difference between two distances is less than $\epsilon$, the two distances (i.e., the two edges) are considered equal; if the difference exceeds $\epsilon$, the edges are considered different. This parameter controls the matching accuracy and is recommended to be set between 0.01 and 0.1. A smaller value increases accuracy but also increases computation time.

(2) Tetrahedron Sorting. Using the filtered sets of equal-length edges, construct undirected graphs to identify all possible tetrahedron combinations. For each possible tetrahedron, store it as six edges, with each edge represented by vertex indices. After constructing all possible tetrahedra, sort them in descending order based on their volumes. When performing matching, prioritize tetrahedra with larger volumes, as this helps improve the stability and accuracy of the matching process.

(3) Tetrahedron Matching. Based on the previous step, if a tetrahedron from each of the two sets has a similar volume—meaning their volume ratio is close to 1 and the relative error is less than $\beta$, then proceed to the next step of matching the edge lengths of these tetrahedra. We set the upper limit for this parameter at 10% to ensure that the volume differences are within an acceptable range. Let the volumes of tetrahedra $T_1$ and $T_2$ be $V_1$ and $V_2$, respectively. The volume matching condition can be expressed as:

$$\left|\frac{V_1}{V_2} - 1\right| < \beta$$

Then, the six edges of the tetrahedra are sorted in descending order by length, and the corresponding edges are compared one by one. If the edge lengths match, further verification of tetrahedron congruence is performed using Singular Value Decomposition (SVD). If the distances between corresponding vertices are all below the threshold δ, we can confirm that the tetrahedra are congruent and consider the match successful. This threshold measures the reliability of the matching result, ensuring that the selected correspondences are within an acceptable error range. If the calculated SVD reconstruction error is less than this threshold, with an upper limit of 0.1, the match is considered valid. This setting is intended to filter out imprecise matches, thereby improving the overall accuracy of the matching results. This process combines multiple validations—volume, edge length, and SVD decomposition—to ensure the precision and robustness of the matching results.

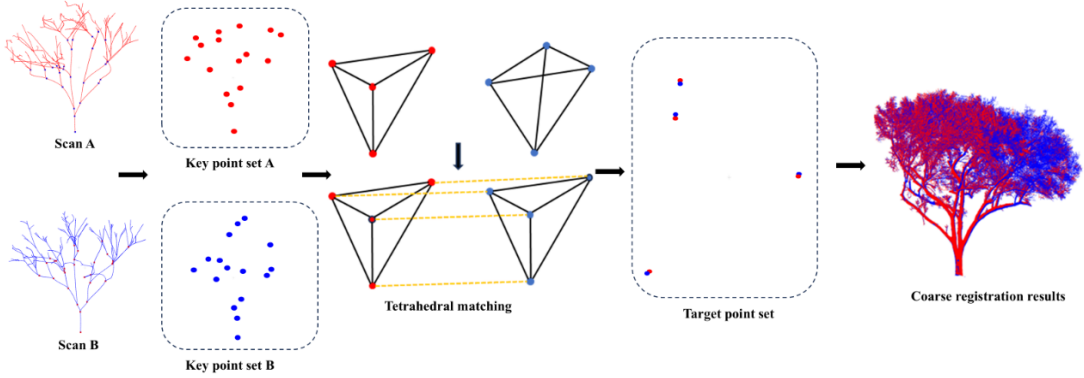

**Fig. 5.** Coarse registration based on tetrahedral point pairs.

*E. Fine Registration*

During the fine registration stage of the single-tree point cloud, due to the complex shape of the trunk, there is often an issue of excessive tight fitting of the branch point clouds. Considering the larger spatial distribution area of the leaf point clouds, and based on experimental testing, the AMRST method uses the leaf point clouds as the base data for estimating the fine registration parameters. Moreover, using only the leaf point clouds reduces the amount of data compared to using the entire tree point cloud, which also reduces the computational load. The fine registration process is shown in Figure 6.

Specifically, based on the coarsely registered leaf point clouds, the LM-ICP (Levenberg-Marquardt Iterative Closest Point) algorithm is used to further refine the registration and estimate the fine registration parameters. The core of the LM-ICP algorithm is to minimize the point-to-point distance between the source point cloud and the target point cloud, which is equivalent to minimizing the cost function shown below. By iteratively optimizing the transformation parameters, the LM-ICP algorithm aims to reach a local minimum of the cost function, ensuring optimal registration between the source and target point clouds.

$$min_{R,t} \sum_{i=1}^{n} |R\, p_i + t - q_{\pi(i)}|^2$$

Here, R and t represent the rotation matrix and translation vector, respectively. $p_i$ is a point in the source point cloud, and $q_{\pi(i)}$ is the nearest corresponding point in the target point cloud.

To optimize the above cost function, the LM-ICP algorithm uses the Levenberg-Marquardt method to update the transformation parameters in each iteration. The parameter update formula is as follows:

$$(J^T J + \lambda D)\, \Delta x = -J^T e$$

Here, $J$ is the Jacobian matrix, $e$ is the error vector, $\Delta x$ is the parameter update vector, $\lambda$ is the damping factor, and $D$ is a diagonal matrix, typically taking the diagonal part of $J^T J$.

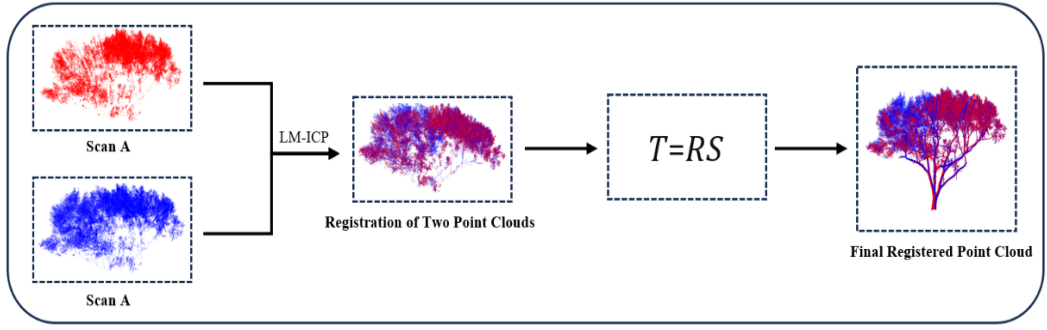

**Fig. 6.** Fine registration based on leaf point clouds

*F. Accuracy Analysis Metrics*

To evaluate the registration accuracy, two parameters were used in the experiment: RMSE and Hausdorff distance. Their calculation formulas are shown below. The Hausdorff distance is a measure of the distance between two point sets, defined as the maximum distance from any point in one set to the nearest point in the other set.

$$\text{RMSE} = \sqrt{\frac{1}{n}\sum_{i=1}^{n}(y_i - \hat{y}_i)^2}$$

Here, nnn is the number of observations, $y_i$ is the $i$ actual observation, and $\hat{y}_i$ is the $i$ predicted value. A smaller RMSE value indicates higher prediction accuracy of the model.

$$H(A,B) = \max\left\{\sup_{a \in A} \inf_{b \in B} d(a,b), \sup_{b \in B} \inf_{a \in A} d(a,b)\right\}$$

Here, $A$ and $B$ are two point sets, and $d(a,b)$ is the distance between points $a$ and $b$. A smaller Hausdorff distance indicates that the shapes or contours of the two point sets are more similar.

### III. RESULTS

*A. Registration Results*

For the two scans of each of the eight trees, wood-leaf separation preprocessing was performed, and tree skeletons were generated, as shown in the figures below. It can be observed that although there are some minor misclassifications due to occlusions, the branch and leaf point clouds are generally well separated. Furthermore, the skeleton generation method used in this paper, despite having some approximations in skeleton details [35], is capable of extracting a tree skeleton that effectively reflects the main structure of the tree, ensuring that the key point set can be constructed successfully.

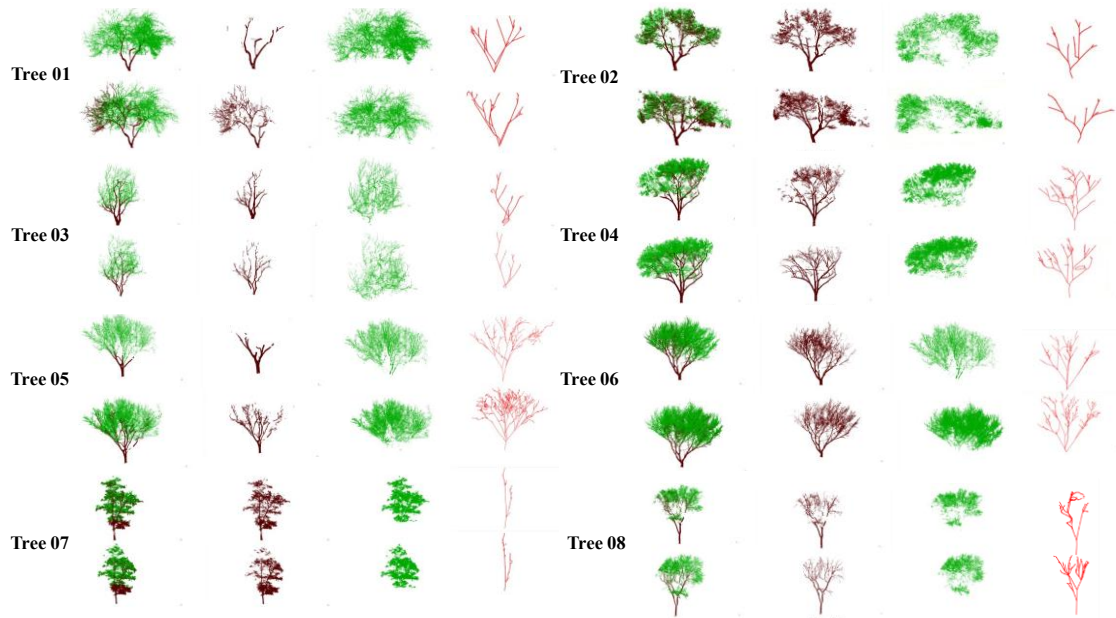

**Fig. 7.** Results of wood-leaf classification. Wood points are brown and leaf points are green.

The coarse and fine registration results of the AMRST method are shown in the figures below. For ease of comparison and analysis, the registration results of the NDT and ICP algorithms are also shown. The first and second columns show the registration results of the NDT and ICP algorithms, respectively, while the third and fourth columns show the coarse and fine registration results of the AMRST method. It is evident from the figures that the registration results obtained using the NDT and ICP algorithms directly are not satisfactory, with ICP performing slightly better than NDT. In contrast, the AMRST method achieved good registration results during coarse registration, and the accuracy was further improved after fine registration.

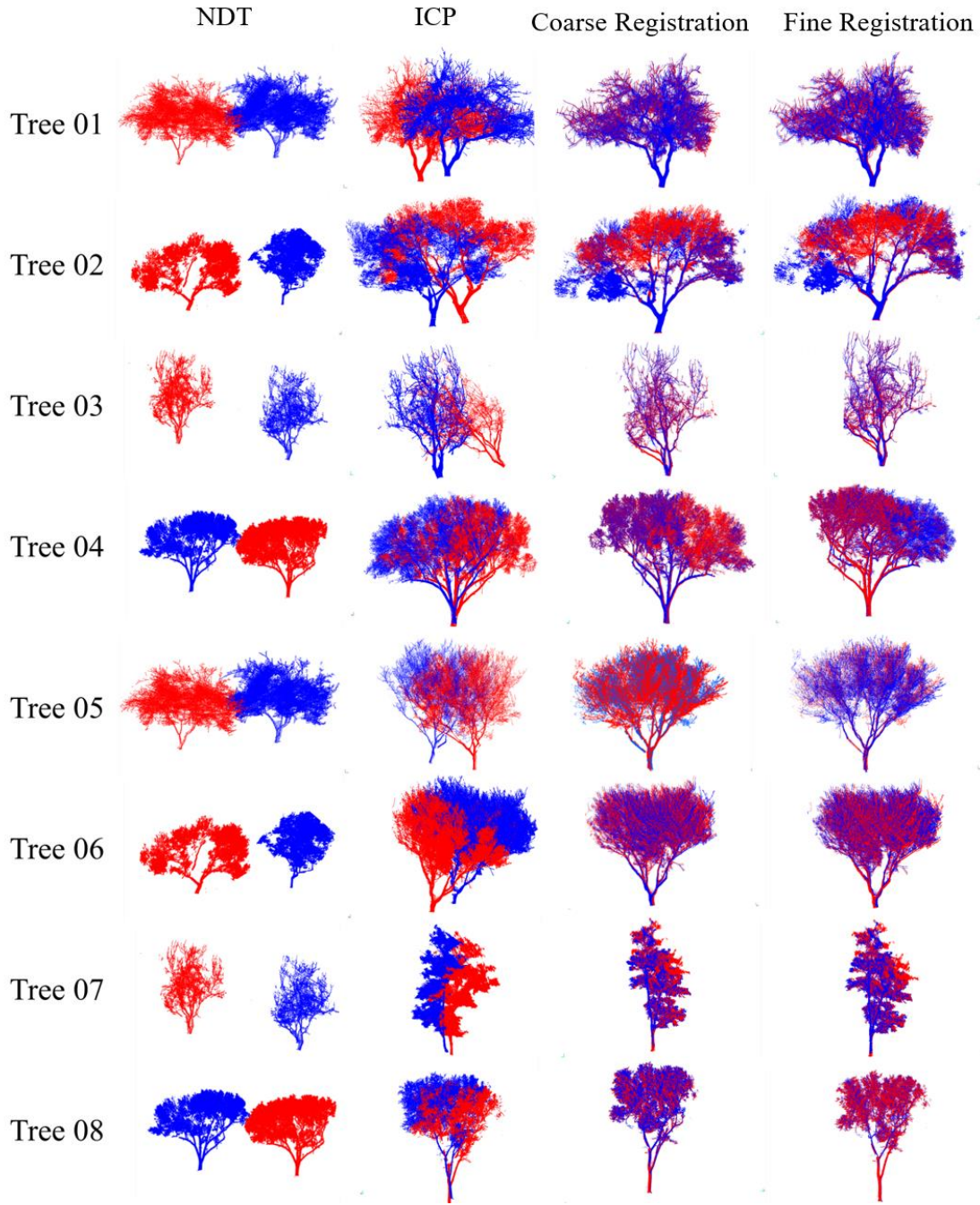

**Fig. 8.** Comparison of the registration results of the NDT method, ICP method and AMRST

*B. Accuracy and Speed Analysis*

This paper evaluates the registration accuracy of the NDT, ICP, and AMRST methods using Root Mean Square Error (RMSE) and Hausdorff

The results show that for all samples, the distance, as shown in Table 2 and Figure 9. The detailed data and line charts are provided for comparison and analysis.

RMSE and Hausdorff distance (HD) of the

AMRST method are significantly lower than those of the NDT and ICP methods. For Tree01 and Tree03 point clouds, the accuracy advantage of the AMRST method is particularly evident. In the coarse registration stage of Tree01, the RMSE of the AMRST method is only 0.08, while the RMSEs of NDT and ICP are 3.83 and 0.65, respectively. In the fine registration stage, the AMRST method further reduces the RMSE to 0.06, and the HD to 0.03, achieving higher registration accuracy. Similarly, for Tree03, the RMSE after coarse registration using the AMRST method is 0.10, which is much lower than 1.03 for ICP and 8.92 for NDT. After fine registration, the RMSE is further reduced to 0.08, and the HD is only 0.05. Clearly, the AMRST method achieves high registration accuracy in both coarse and fine registration for single-tree point cloud data.

TABLE II

RMSE AND HAUSDORFF DISTANCE OF REGISTRATION RESULTS (UNIT: METERS)

| | NDT | | ICP | | AMRST | | | |
| | | | | | Coarse Registration | | Fine Registration | |
| | RMSE | HD | RMSE | HD | RMSE | HD | RMSE | HD |
| Tree01 | 3.83 | 6.83 | 0.65 | 3.05 | 0.08 | 0.77 | 0.06 | 0.03 |
| Tree02 | 9.83 | 26.10 | 3.25 | 8.72 | 2.25 | 6.83 | 1.56 | 1.63 |
| Tree03 | 8.92 | 13.33 | 1.03 | 3.07 | 0.11 | 2.60 | 0.08 | 0.05 |
| Tree04 | 8.71 | 18.75 | 0.84 | 5.72 | 0.19 | 3.38 | 0.07 | 1.13 |
| Tree05 | 11.98 | 16.33 | 0.54 | 2.66 | 0.15 | 1.29 | 0.15 | 1.27 |
| Tree06 | 2.63 | 14.37 | 0.67 | 2.95 | 0.08 | 1.00 | 0.07 | 0.91 |
| Tree07 | 1.11 | 4.30 | 0.88 | 2.65 | 0.15 | 1.42 | 0.10 | 1.41 |
| Tree08 | 18.22 | 9.36 | 0.61 | 4.87 | 0.15 | 1.11 | 0.15 | 0.89 |

It can also be seen from Figure 9 that the AMRST method demonstrates significant advantages in registration accuracy. In particular, Figure 9a shows the comparison of RMSE values for different methods across the samples, while Figure 9b shows the comparison of Hausdorff distances. Compared to the NDT and ICP methods, AMRST achieves high accuracy in the coarse registration stage, with lower RMSE and Hausdorff distances, effectively achieving high-quality initial registration. In the fine registration stage, AMRST further optimizes the registration results, significantly reducing errors and achieving a higher overall registration accuracy.

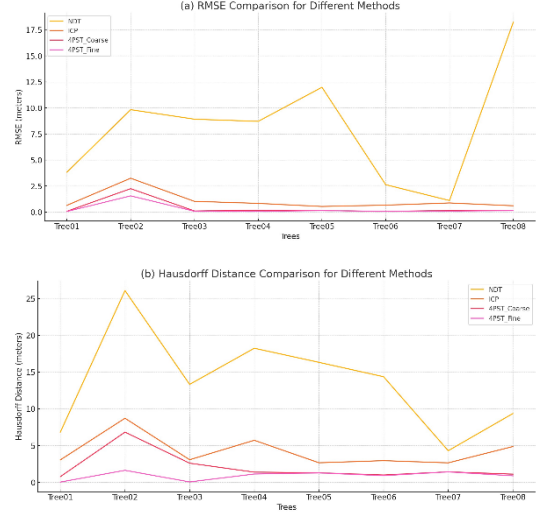

**Fig. 9.** Line chart of registration accuracy of three methods. a. lines for RMSE, b. Lines for Hausdorff distance.

In addition to achieving good registration accuracy, registration efficiency is also an important metric. Table 3 shows the time taken to register the eight trees using the AMRST, NDT, and ICP methods, with specific details for the coarse registration time, fine registration time, and total time for the AMRST method.

TABLE III

TIME COST OF THREE METHODS (UNIT: SECONDS)

| | NDT | ICP | AMRST | |
| | | | Coarse Registration | Fine Registration |
| Tree01 | 86.95 | 166.82 | 0.73 | 0.13 |
| Tree02 | 230.55 | 1211.35 | 1.14 | 0.90 |
| Tree03 | 9.85 | 209.04 | 0.36 | 0.42 |
| Tree04 | 47.70 | 466.68 | 1.46 | 0.09 |
| Tree05 | 14.48 | 99.41 | 0.86 | 0.86 |
| Tree06 | 115.09 | 366.54 | 7.98 | 0.16 |
| Tree07 | 79.70 | 325.62 | 0.33 | 0.15 |
| Tree08 | 13.78 | 93.30 | 1.14 | 0.10 |

From Table 3, it can be seen that the ICP method takes the longest registration time among the three methods. The NDT method is faster than the ICP method but still consumes a considerable amount of registration time, especially when the point cloud size is large. For instance, ICP takes 1211.35 seconds to register the point cloud of Tree02. In contrast, the AMRST method's registration time is significantly lower than that of the other two methods, making it the most efficient of the three. Specifically, when registering Tree01, the total time taken by the AMRST method is 0.86 seconds, whereas the NDT and ICP methods take 86.95 seconds and 166.82 seconds, respectively, which are 101 times and 193 times longer than the AMRST method. Similarly, for Tree04, the AMRST method's registration time is 1.54 seconds, while the NDT and ICP methods take 47.70 seconds and 466.68 seconds, respectively, which are 31 times and 303 times longer than the AMRST method. Clearly, the AMRST method shows a significant advantage in computational efficiency when registering complex single-tree point clouds, effectively reducing the time cost during the registration process.

## IV. DISCUSSION

The experimental results above show that the AMRST method achieves high-quality coarse registration by constructing a key point set using tree skeletons and finding corresponding vertices of similar tetrahedra. It then enhances the registration precision by refining the process with leaf point clouds. The use of tree skeletons and key point extraction significantly reduces the data size during the preprocessing stage due to leaf separation. In the experiments involving point cloud registration of eight trees from different species, the AMRST method consistently outperforms traditional NDT and ICP methods in terms of both precision and computational efficiency, demonstrating its excellent performance and robustness. Therefore, the AMRST method is an effective solution for automatic point cloud registration that leverages the spatial consistency of tree structures.

The AMRST method is divided into two main stages: coarse registration and fine registration. In both stages, a portion of the data from the leaf separation process is used for computation. Coarse registration mainly uses the stem point cloud to generate the tree skeleton, which greatly reduces the computational load and significantly improves efficiency. Moreover, the coarse registration is not overly sensitive to the accuracy of the tree skeleton, which ensures good registration precision at this stage. In the fine registration stage, leaf point clouds are used, further reducing the data for registration. Although the method uses an improved version of ICP that requires iteration, the high accuracy of the coarse registration provides excellent starting conditions, allowing the fine registration to converge quickly and achieve precision improvement in a short amount of time.

However, since the AMRST method relies on tree skeleton generation and key point extraction during coarse registration, it may struggle with trees whose branch structure is not clearly defined, as this could hinder key point extraction and affect registration performance. Additionally, the density of the tree's leaves can influence the registration process. When the leaves are too dense, it may obscure a large portion of the branches, reducing the number of recognizable and matchable key points, thereby increasing the difficulty of registration. Future research could explore strategies for achieving accurate and effective registration in situations where branch features are not distinct or when heavy leaf occlusion occurs, further improving the method's applicability and robustness.

Ⅴ. CONCLUSION

In this study, a novel method for unlabeled automatic registration of single-tree point clouds based on similar tetrahedra is proposed for the application of ground-based LiDAR technology in forestry surveys. The method utilizes branch point clouds and leaf point clouds, obtained through leaf-stem separation, to assist with coarse and fine registration, respectively. During the registration process, measures such as leaf-stem separation, tree skeleton generation, and extraction of similar tetrahedral key point pairs are employed, which significantly enhance the accuracy, efficiency, and robustness of single-tree point cloud registration.Complex tree structures and environmental conditions can affect the registration performance of the AMRST method. Therefore, future research will focus on improving the method's applicability in more challenging environmental scenarios. Looking ahead, this method is expected to play an important role in forestry surveys and tree point cloud data processing, providing technical support for the precise measurement and management of forestry resources.